\documentclass[letterpaper, 10 pt, journal, twoside]{IEEEtran}

\IEEEoverridecommandlockouts
\usepackage{amsmath, amssymb}

\usepackage{graphics} 
\usepackage{epsfig} 
\usepackage{times} 
\usepackage{amsmath} 
\usepackage{amssymb}  
\usepackage[table]{xcolor}
\usepackage[ruled,linesnumbered, noend]{algorithm2e}
\SetKwRepeat{Do}{do}{while}
\usepackage{adjustbox}
\usepackage{wrapfig}
\usepackage{siunitx}

\usepackage{soul}
\usepackage{booktabs}

\usepackage{cite}
\usepackage{comment}
\usepackage{multirow}
\usepackage{graphicx}
\usepackage{wrapfig}
\usepackage{balance}
\usepackage{subfigure}
\usepackage{capt-of}

\newcommand{\ve}[1]{\mathbf{#1}} 
\newcommand{\tve}[1]{\tilde{\mathbf{#1}}} 

\usepackage{color, colortbl}
\definecolor{Gray}{gray}{0.9}

\usepackage{pifont}
\title{Human Arm Pose Estimation with a Shoulder-worn Force-Myography Device for Human-Robot Interaction}

\author{Rotem Atari, Eran Bamani and  Avishai Sintov
\thanks{This work was supported by the Israel Science Foundation (grant No. 1565/20).}
\thanks{R. Atari, E. Bamani and  A. Sintov are with the School of Mechanical Engineering, Tel-Aviv University, Israel. Corresponding Author: sintov1@tauex.tau.ac.il.}
}

\begin{document}

\maketitle

\begin{abstract}
Accurate human pose estimation is essential for effective Human-Robot Interaction (HRI). By observing a user's arm movements, robots can respond appropriately, whether it's providing assistance or avoiding collisions. While visual perception offers potential for human pose estimation, it can be hindered by factors like poor lighting or occlusions. Additionally, wearable inertial sensors, though useful, require frequent calibration as they do not provide absolute position information. Force-myography (FMG) is an alternative approach where muscle perturbations are externally measured. It has been used to observe finger movements, but its application to full arm state estimation is unexplored. In this letter, we investigate the use of a wearable FMG device that can observe the state of the human arm for real-time applications of HRI. We propose a Transformer-based model to map FMG measurements from the shoulder of the user to the physical pose of the arm. The model is also shown to be transferable to other users with limited decline in accuracy. Through real-world experiments with a robotic arm, we demonstrate collision avoidance without relying on visual perception.
\end{abstract}


\section{Introduction}
\label{sec:introduction}

Human-Robot Interaction (HRI) within a shared workspace involves either coexistence or collaborative interaction. In coexistence scenarios, humans and robots work independently on distinct tasks while avoiding each other \cite{schiavi2009}. In collaborative interaction, they interact dynamically to accomplish a shared goal \cite{De2012}. Both scenarios require the robot to accurately understand and anticipate the human's intentions and arm movements in real-time. Applications can  include medical procedures \cite{Okamura2010}, rehabilitation \cite{Mohebbi2020}, factory assistance \cite{Matheson2019} and domestic robotics \cite{Qin2023}. In these scenarios, the robot's ability to observe human motion and respond is crucial for effective interaction.

We consider the problem of estimating the current pose of the entire human arm. Estimation models are usually based on either visual perception or wearable devices. The most notable approach for the former is the use of human pose estimation models (i.e., \textit{Skeleton} models \cite{Osokin2018,Lugaresi2019}). Such a model locates key points (e.g., head, shoulders, elbows, wrists, hips and knees) on a human body within an image or video, that can be used to reconstruct a person's pose in 2D or 3D space. Hence, a camera can observe the user, estimate arm poses in real-time, and proactively initiate actions to avoid collisions or plan task completion \cite{Schmidt2014}. However, 
the sole reliance on continuous visual feedback can hinder task performance in scenarios with visual uncertainty, such as poor lighting, occlusions, long distance and multiple users in the scene. Additionally, visual sensing demands substantial data and computational resources \cite{Toshev2014}, potentially limiting its practicality in certain applications.

\begin{figure}
    \centering
    \includegraphics[width=\linewidth]{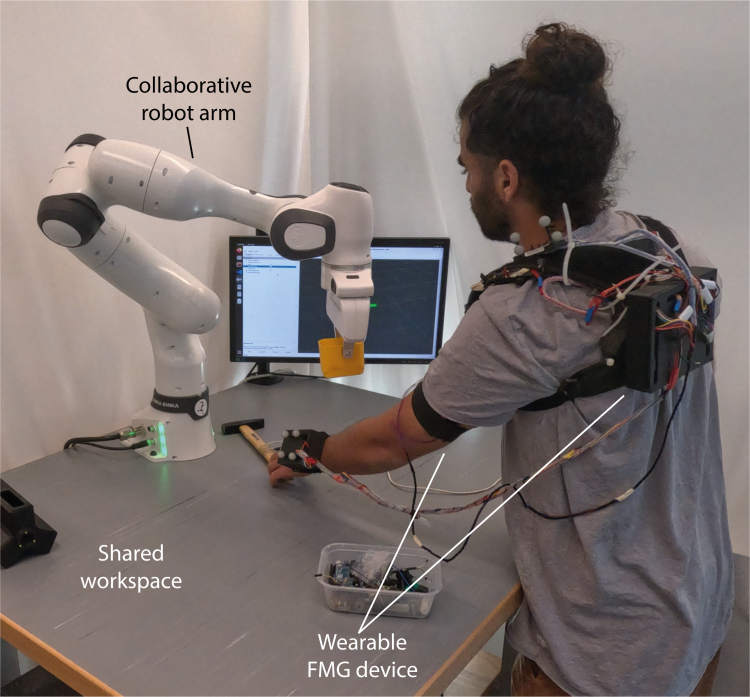}
    \caption{A user is working in a share workspace with a robotic arm. A wearable Force-Myography (FMG) device is used to estimate the pose of the human arm in real time. In this example, when the user reaches to pick up a tool, the robot halts its motion to avoid interference and potential collisions.}
    \label{fig:Cover}
\end{figure}
\begin{figure*}
\vspace{1cm}
    \centering
    \includegraphics[width=\linewidth]{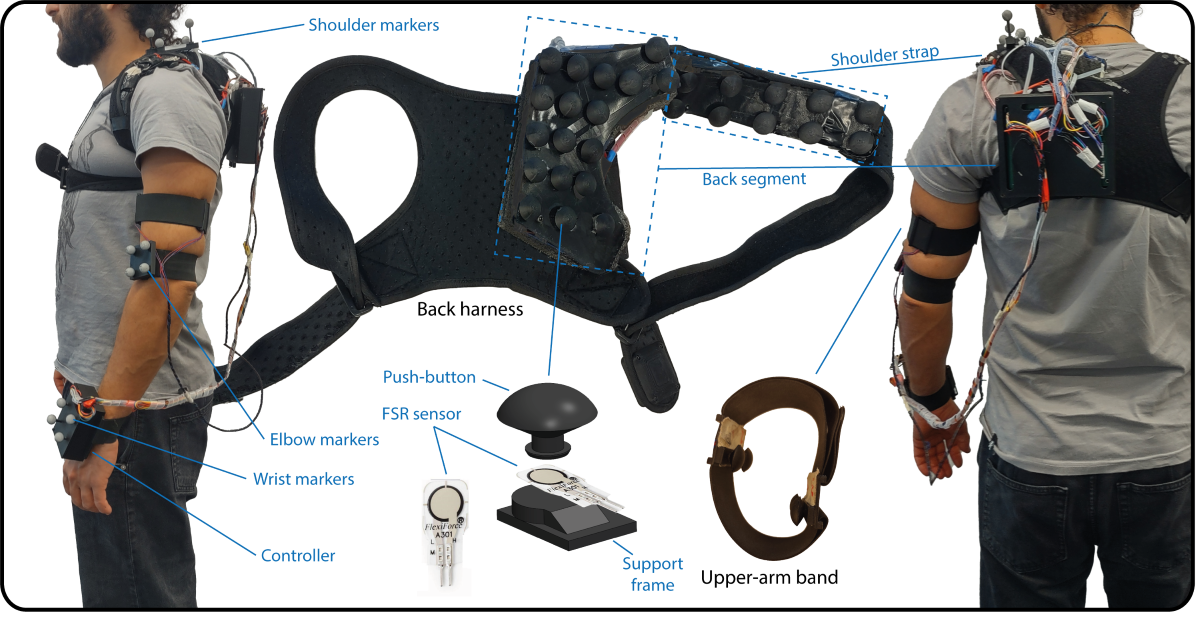}
    \caption{The wearable FMG device includes a back harness and an upper arm band, with a total of 32 FSR sensors. Reflective markers are fixed on the shoulder, elbow and wrist for data collection.}
    \label{fig:intro}
    \vspace{-0.2cm}
\end{figure*}

To cope with the vision limitations, wearable technology has been proposed with sensing modalities such as Electro-Myography (EMG) \cite{Wang2024} and Ultrasound \cite{Niu2024}. In general, estimating arm poses solely through wearable sensing can enhance the robustness of HRI systems, particularly in challenging environments with limited or occluded visual information. Later, potential fusing of wearable sensor data with visual inputs can achieve more accurate and reliable pose estimation. The prominent approach is the use of body-worn Inertial Measurement Units (IMU) \cite{Prayudi2012}. Typically, one or two IMU sensors are placed on the user's arm and used to estimate arm pose \cite{Yun2006,Atrsaei2018,García-de-Villa2023}. IMUs typically combine accelerometers, gyroscopes and magnetometers, and can be used to estimate kinematic data such as velocity, position and orientation. These estimations are often jeopardized by noise and drift. Kalman filtering and machine learning techniques are commonly employed to address these challenges and improve the accuracy of the estimated motion \cite{Dumpis2022}. However, IMU-based estimations require frequent calibration as they provide relative, rather than absolute, measurements \cite{DeVrio2023}.  
With similar calibration requirements, IMUs were also used to predict the future target of the users arm for seamless HRI \cite{Cui2022,kahanowich2024}.

Force-Myography (FMG) is a non-invasive technique for measuring muscle activity \cite{Amft2006}. FMG involves placing simple force sensors on the skin to monitor muscular contractions and relaxations. Compared to EMG, FMG is known for its ease of acquisition, high accuracy and robustness to positioning variations \cite{Jiang2017,Belyea2019}. FMG offers a better signal-to-noise ratio and anti-interference ability, providing more stable signals with resistance to interference such as sweating and electromagnetic disturbances \cite{BiosensorsReview}. In addition, FMG technology utilizes simple force-sensitive resistors, resulting in a low-cost device. These advantages have led to its successful application in rehabilitation studies \cite{Yap2016}, tele-operation \cite{Mizrahi2024}, prosthetics \cite{radmand2016,cho2016force}, and classification of hand gestures \cite{Gantenbein2023} and held objects \cite{kahanowich2021robust,Bamani2022}. In all of these applications, FMG sensing was conducted on the forearm of the user and, thus, usually provides only information regarding the state of the hand and fingers. To the best of the authors' knowledge, FMG sensing has not been previously applied to model the complete state of the human arm, and specifically by measuring perturbations of shoulder muscles.

In this letter, we explore the feasibility of utilizing FMG for comprehensive modeling of the human arm, with the goal of natural HRI. In a novel approach, FMG sensors are strategically placed on the upper back, shoulder and arm, enabling the acquisition of data that can be used to infer arm pose. This is the first application of FMG on shoulder muscles and for modeling the state of the entire human arm. To estimate the instantaneous pose, we propose a novel approach utilizing the Transformer architecture \cite{Zerveas2021}, leveraging its ability to process temporal sequential data effectively. We then demonstrate its use in an HRI scenario where a robot must perform its own task without interfering or colliding with the human user. Unlike the relative positioning of IMUs, FMG offers absolute positioning with respect to the human torso without the need for constant calibration. Also, in contrast to visual perception, it is environment-agnostic and does not rely on line-of-sight or ambient lighting conditions. Hence, it can be integrated in the future to the clothing of users, enabling seamless sensing and data collection in various applications. 

This work pioneers the use of FMG for full-arm pose estimation, enabling robust human-robot interaction. Key contributions include:
\begin{itemize}
    \item \textit{Novel FMG Application}: We introduce a novel approach to FMG, positioning sensors on the shoulder to capture arm pose information.
    \item \textit{Transformer-based Model}: We propose a Transformer-based model to effectively map FMG signals to accurate arm pose estimates.
    \item \textit{Cross-User Generalization}: We demonstrate the model's ability to generalize to new users, even with varying body dimensions.
    \item \textit{Real-world Validation}: The model's effectiveness is validated in real-world HRI scenarios with a collaborative robot arm.
\end{itemize}
While our focus is on robotic arm collaboration, the proposed approach can also be adapted for applications involving prosthetic hands, drones, teleoperation, and virtual reality.

\section{Method}
\label{sec:System}

\subsection{Problem Statement}

We consider a shared workspace denoted by $\mathcal{W} \in \mathbb{R}^3$. It is assumed that the body pose of the human user in front of $\mathcal{W}$ is known and relatively static. The general problem involves enabling a robot to estimate the pose of a human arm within the shared workspace, with the aim of facilitating collaborative tasks or avoiding collisions. The state of the user's arm is represented by the tuple $\ve{v}=\{\ve{p}_{el},\ve{p}_{wr}\}\in\mathcal{P}$ where $\mathcal{P}\subset \mathbb{R}^3\times\mathbb{R}^3$ and is composed of the spatial positions $\ve{p}_{el}\in\mathbb{R}^3$ and $\ve{p}_{wr}\in\mathbb{R}^3$ of the elbow and wrist, respectively. The positions are measured with respect to the position of the shoulder $\ve{p}_{sh}\in\mathbb{R}^3$. Furthermore, an FMG device is positioned on the shoulder and upper arm of the user. Hence, let $\ve{x} \in \mathbb{R}^{n}$ represent the state of the musculoskeletal system, which is captured through the $n$ FMG signals acquired from the FMG device. We search for a mapping $\Gamma: \mathcal{C}  \to \mathcal{P}$ where $\mathcal{C}$ is some product of the FMG space in $\mathbb{R}^{n}$. Map $\Gamma$ should approximate the state of the human arm with respect to the body based on the FMG data.

\subsection{FMG System}

The proposed FMG device is based on a back harness and an upper arm band with a total of $n=32$ Force-Sensitive Resistors (FSR) model FlexiForce A301. FSR sensors are polymer films that exhibit changes in electrical resistance when subjected to varying pressure. Each FSR sensor is equipped with a push-button mechanism. The button has a spherical design to allow better attachment onto the muscle and a larger surface area, maximizing the amount of information extracted from each muscle group. This adaptability accommodates the variations in the user's skin even when the surface is uneven. All FSR sensors are connected to an Arduino Uno micro-controller through a voltage divider of 150k$\Omega$. This configuration enables real-time reading of the sensors at a maximum stable frequency of 100 Hz, yielding similar sampling frequency.

The back harness is comprised of 28 FSR and is tightened to the back using shoulder straps. The FSR sensors on the harness's dorsal cover the upper back of the user at the shoulder of the measured arm as seen in Figure \ref{fig:intro}. The back segment of the dorsal covers, with 18 FSR sensors, the Infraspinatus muscle and the back side of the Trapezius muscle \cite{Terry2020}. In addition, 10 FSR sensors are aligned on the shoulder strap and sense the top of the Trapezius muscle. The separated band with four sensors is wrapped around the upper arm and sense perturbations of the triceps with four FSRs. All FSR sensors are distributed in relatively equal spacing and in tabular formations within each section.

To collect labeled data, a motion capture (MoCap) system was used to measure  $\ve{v}$. Therefore, reflective markers were positioned on the shoulder, elbow and wrist of the measured arm. They were measured with respect to the coordinate frame of $\mathcal{W}$ determined in the calibration of the MoCap. Then, the positions of the elbow and wrist with respect to the shoulder can be computed by simple subtraction, making them independent of any environment. Consequently, an FMG measurement $\ve{x}_i$ can be labeled by an arm state $\ve{v}_i$. 


\subsection{Data Collection}
\label{sec:data_collection}

Train and test datasets were collected by synchronously recording FMG signals with the three key points on the human arm. This research investigates the feasibility and requirements of training a $\Gamma$ model using data from only a single participant and potentially transferring it to novel users. To achieve this, data was collected from a one participant across $K$ separate sessions and $M$ samples per session. In each session, the device was taken off and re-worn to include variations in sensor placement and tightening forces. 

During each session, the participant was instructed to perform a diverse set of arm movements within the front workspace. These movements included scanning through the space with varied arm postures and velocities, flexing and extending the arm at different elbow positions and velocities, reaching to different target locations, and holding a range of static postures. Hence, the participant comprehensively explores the full arm range of motion, ensuring that both high and low velocity movements were captured across a wide array of configurations. This approach allows the dataset to accurately represent fine muscle movements as well as gross motor actions. 
The resulting dataset is a set of $N = KM$ labeled FMG measurements $\mathcal{Q} = \{(\ve{x}_i,\ve{v}_i)\}_{i=1}^N$. A similar test dataset was collected in independent sessions for evaluating trained models. 



\subsection{Pose Estimation Model}

We now aim to train a data-based model $\Gamma$ to acquire pose estimation based on FMG data. 
We hypothesize that a temporal analysis of FMG signals can enhance arm pose estimation. This assumption is grounded in the belief that FMG signals vary with arm velocities, and temporal sequences can effectively embed these variations. We define the space of temporal FMG sequences by $\mathcal{C}_H\subset\mathbb{R}^n\times\ldots\times\mathbb{R}^n$ where $H$ is the length of the sequence. Therefore, sequential batches of length $H$ are extracted from dataset $\mathcal{Q}$ and labeled with the corresponding pose based on the last signal in the sequence. Let $\ve{x}(t)$ and $\ve{v}(t)$ be the FMG and arm states, respectively, at time $t$. Hence, an FMG sequence
\begin{equation}
    \ve{c}(t)=\{\ve{x}(t-H),\ldots,\ve{x}(t)\}\in\mathcal{C}_H
\end{equation}
corresponding to time $t$ is labeled by $\ve{v}(t)$. The sequences are sampled  using a sliding window over the episodes in $\mathcal{Q}$ with step size $H/2$ to ensure diversity. This step size assists in reducing data redundancy while maintaining the temporal structure. Similar sampling is done to the test set but with a window step size of 1. The pre-processing step yields a modified dataset $\mathcal{Q}' = \{(\ve{c}_i,\ve{v}_i\}_{i=1}^W$ where the number of sequences $W$ depends on $H$ and $N$. Consequently, we search for a map $\Gamma_\theta:\mathcal{C}_H\to\mathcal{P}$ where $H$ is an hyper-parameter to be optimized and $\theta$ is the trainable weight vector of the model.

\begin{figure*}
    \centering
    \includegraphics[width=0.85\linewidth]{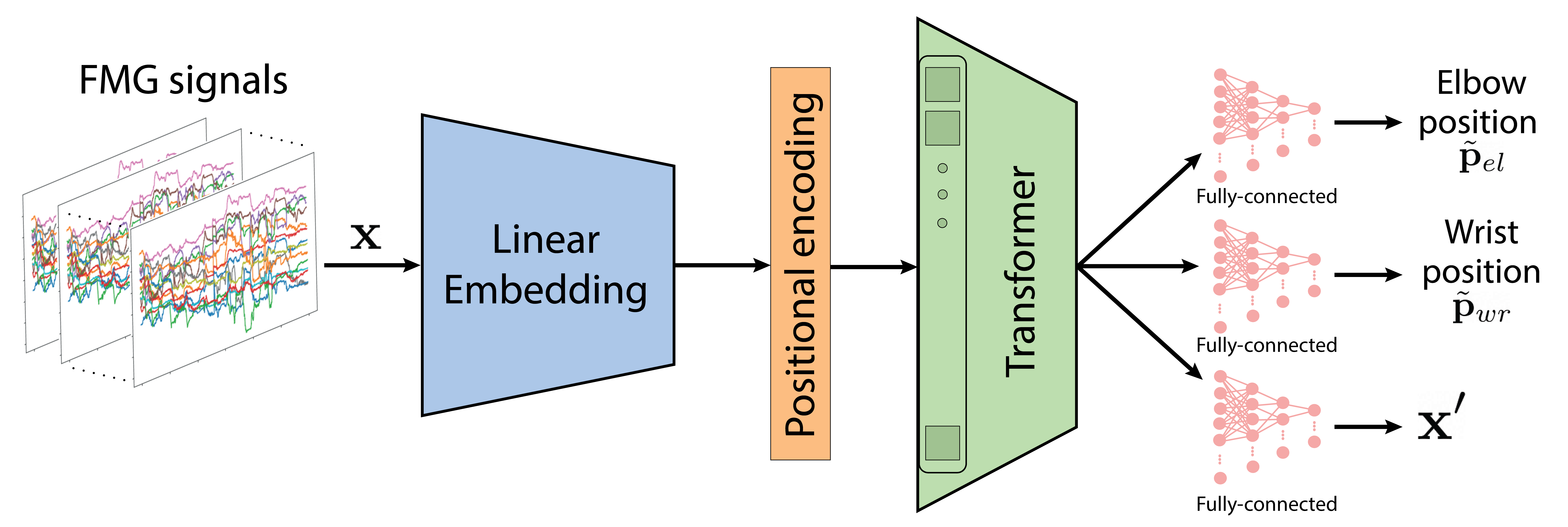}
    \vspace{-0.5cm}
    \caption{Illustration of the Transformer-based model for mapping temporal FMG signals to the positions of the elbow and wrist.}
    \label{fig:model}
    \vspace{-1.cm}
\end{figure*}

Mapping $\Gamma_\theta$ can be formulated as a multivariate time series representation learning problem, where the goal is to map a temporal FMG signal $\ve{c}(t)$ to the arm pose $\ve{v}(t)$. While sequential modeling can be done using models such as Recurrent Neural Networks (RNN) \cite{Anvaripour2020} or Convolutional Neural Networks (CNN), Transformers have an emerged in the last few years as a powerful and data efficient tool for modeling sequential data such as language and time series \cite{Zerveas2021}. The success of the Transformer can be attributed to its ability to model long-range dependencies using a multi-head attention mechanism, which allows it to attend to different time steps simultaneously. Unlike RNN-based models that process data sequentially, Transformers process all input data at once, making them well-suited for tasks where long-term dependencies are important. This capability is particularly beneficial for FMG data, which requires learning complex dependencies between muscle activity and arm movements. Therefore, we base our  $\Gamma_\theta$ model on the Transformer Encoder architecture and later compare it to other existing data-based models.

Our model, illustrated in Figure \ref{fig:model} consists of an embedding layer that maps the input data into a space of dimension $d_{model}$. A positional encoding layer is applied to inject information about the sequence order into the model. The model includes a transformer encoder with multiple layers, each comprising a multi-head attention mechanism, a feed-forward network with $d_{ff}$ hidden layers and ReLU activation after each layer. Dropout is applied after each attention and feed-forward sub-layer. The output of the Transformer is passed through a series of Fully Connected (FC) layers, each reducing the dimensionality of the feature space, yielding the final position estimation outputs $\tve{p}_{el}$ and $\tve{p}_{wr}$. In addition, the output of the Transformer is passed through another fully-connected network outputting $\ve{x}'$, the reconstruction of the FMG input $\ve{x}$. The Transformer-based model is trained in two phases \cite{Zerveas2021}. In the first phase, the model is trained in an unsupervised manner where the positional outputs are ignored and the reconstruction loss $\mathcal{L}=\|\ve{x}-\ve{x}'\|$ 
is minimized. In the second phase, supervised learning is employed to minimize the loss given by
\begin{equation}
    \mathcal{L}=\frac{1}{J}\sum_{i}^J \{(\tve{p}_{el,i}-\ve{p}_{el,i})^2 + (\tve{p}_{wr,i}-\ve{p}_{wr,i})^2\}  
\end{equation}
where $\tve{p}$ and $\ve{p}$ are the predicted and ground-truth positions, respectively, and $J$ is the number of samples in the batch. 




\section{Experiments}
\label{sec:experiments}

In this section, we evaluate the FMG-based arm pose estimation model and the ability of the model to be used in a shared workspace within an HRI scenario. The data collection and experiments were conducted with the approval of the ethics committee at Tel-Aviv University under application No. 0007829. All models were trained with an NVIDIA GeForce RTX 3060 Ti GPU. Hyperparameter tuning was performed using Ray-Tune \cite{Liaw2018}, optimizing the model parameters across multiple configurations. Videos of data collection and experiments can be seen in the supplementary material.


\subsection{Dataset}

Data was collected as described in Section \ref{sec:data_collection} on a single human subject. The user contributed data across $K=48$ sessions with $M=20,000$ average samples per session. Between each session, the FMG device was removed and re-positioned. In addition, the participant's torso was kept in an upright, fixed position to minimize torso movement, thus isolating the analysis to arm movements. The collection yielded a total of approximately $N=10^6$ samples in $\mathcal{Q}$. In addition, a separate and independent test set was collected with 40,000 samples over two sessions. Then, the dataset was processed into temporal series with horizon $H=128$ leading to $W=7,306$ temporal sequences in $\mathcal{Q}'$. The value for $H$ was chosen based on hyper-parameter optimization. A similar test set was generated with 19,371 labeled sequences.


\begin{table}[ht]
\centering
\caption{Evaluation Metrics for Different Models}
\vspace{-0.3cm}
\label{tab:evaluation}
\begin{tabular}{lccc}
\toprule
\multirow{2}{*}{Model} & \multicolumn{2}{c}{Position error (mm)} & Inference  \\\cmidrule{2-3}
& Elbow & Wrist & time (ms) \\
\midrule
FC-NN           & 73 $\pm$ 60 & 143 $\pm$ 33 & 0.37 \\
1DCNN           & 71 $\pm$ 55 & 140 $\pm$ 55 & 0.27 \\
DLinear         & 81 $\pm$ 80 & 153 $\pm$ 80 & 5.7  \\
CNN-LSTM        & 67 $\pm$ 65 & 138 $\pm$ 65 & 0.88 \\
Transformer     & \cellcolor[HTML]{C0C0C0}60 $\pm$ 60 & \cellcolor[HTML]{C0C0C0}120 $\pm$ 33 & 1.3  \\
\bottomrule
\end{tabular}%
\vspace{-0.5cm}
\end{table}
\begin{figure*}[h]
    \vspace{1cm}
    \centering
    \includegraphics[width=\linewidth]{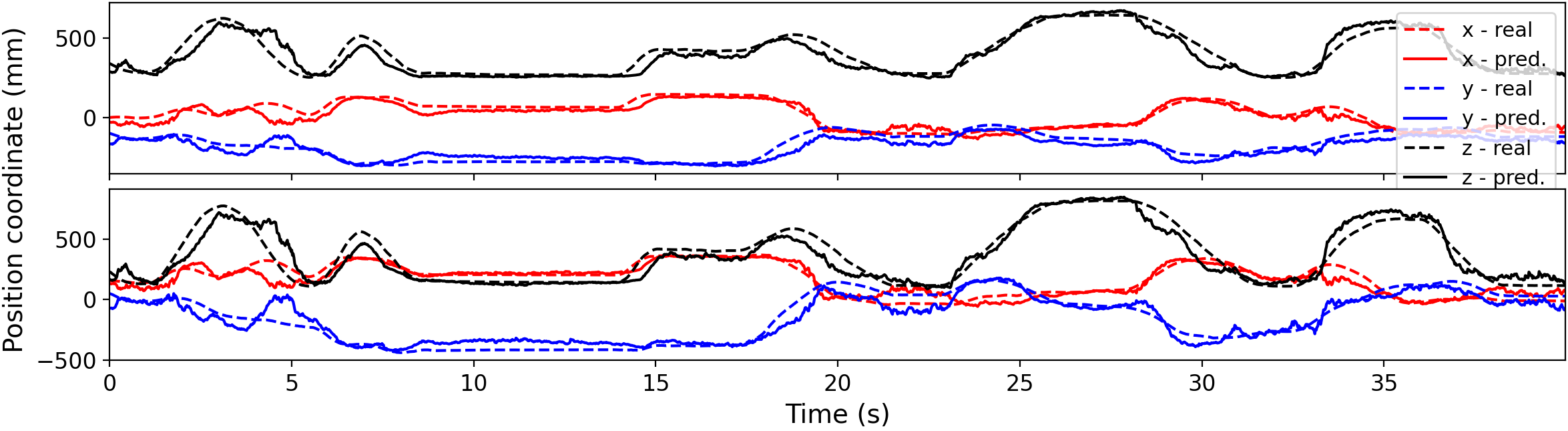}
    \vspace{-0.8cm}
    \caption{Real and predicted positions of the (top) elbow and (bottom) wrist, with regard to motion time. The mean position errors for the elbow and wrist along the example path are 65.4 mm and 116.6 mm, respectively.}
    \label{fig:xyz}
    \vspace{-0.5cm}
\end{figure*}
\begin{figure}
    \centering
    \includegraphics[width=\linewidth]{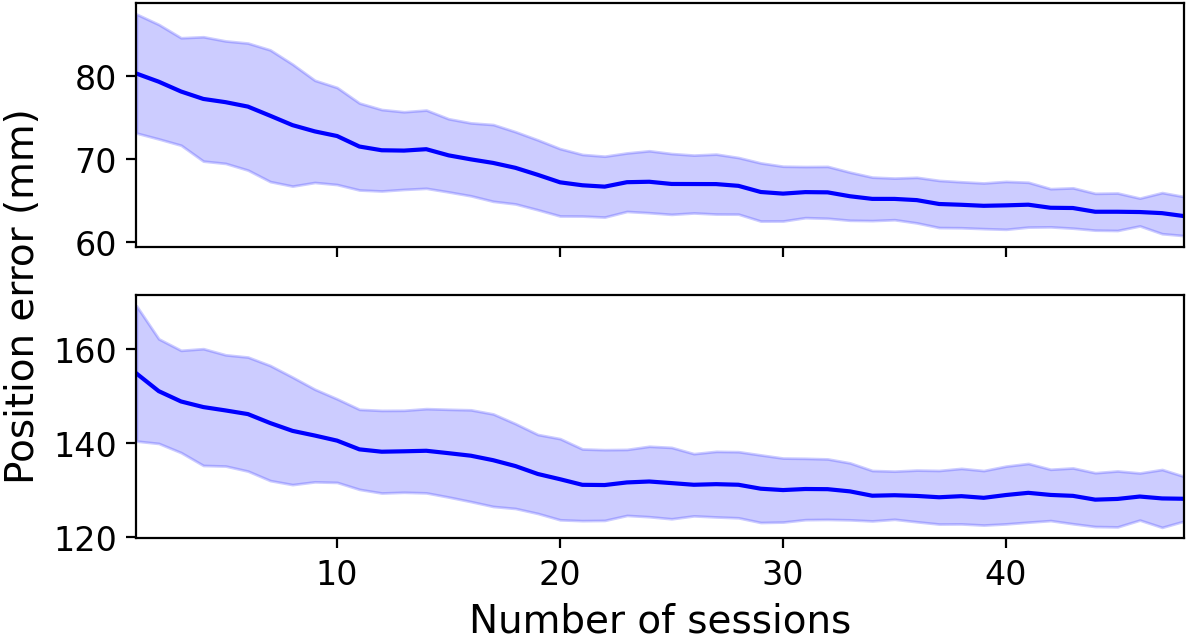}
    \vspace{-0.6cm}
    \caption{Position errors of the (top) elbow and (bottom) wrist, with regard to the number of recorded sessions used for training the Transformer-based model. The results show mean and standard deviation values for 15 training attempts while sampling different sessions in each.}
    \label{fig:error_data}
    \vspace{-0.7cm}
\end{figure}

\subsection{Model Evaluation}

In this section, we analyse the Transformer-based model. First, we compare our proposed model to other existing models including Fully-Connected Neural Network (FC-NN), 1DCNN \cite{Kiranyaz2021}, DLinear \cite{Zeng2023} and Long Short-Term Memory with CNN (CNN-LSTM) \cite{Sainath2015}. The FC-NN is the baseline where the temporal data is flattened to a $n\times H$ vector and fed into a series of fully-connected layers. 1DCNN is a type of CNN specifically designed for processing sequential data, such as text or time series. It applies filters to input sequences to extract relevant features and learn patterns. 
DLinear is a deep linear model designed for forecasting long-term time series and combines the strengths of linear models with the flexibility of deep architectures. It uses a linear transformation to capture temporal dependencies while maintaining computational efficiency. Lastly, CNN-LSTM combines CNNs for feature extraction and LSTM for sequence processing. It passes the data through a series of 1D convolutional layers followed by a sequence of LSTM layers and an FC layer. 

The hyper-parameters of all models were optimized to provide a minimal loss solution over the test set. Specifically for our Transformer-based model, it is trained with a learning rate of 0.001, a weight decay of $\num{5e-5}$ and batch size of 32. Furthermore, the embedding of the transformer maps into $d_{model}=32$ dimension space following two layers, each with eight attention heads and a feed-forward network of $d_{ff}=128$ hidden layers. Finally, the Transformer's output is fed into three fully connected networks, each with two hidden layers of size [32, 10] and a dropout rate of $0.2$.

Table \ref{tab:evaluation} summarizes the comparative analysis between the models. The results show the Root Mean Square Error (RMSE) for the elbow and wrist positions over the test data and the mean inference time for a single prediction. For both elbow and wrist, the Transformer provides predictions with the lowest error. While the elbow estimation is influenced only by the motions of the shoulder, the wrist’s position is more affected by both the elbow and shoulder angles and, thus, is more susceptible to errors propagated through the kinematic chain. This is reflected in the larger error values observed for wrist position estimation compared to the elbow estimations.
While the Transformer architecture offers better performance compared to other models, it generally requires slightly longer inference times due to its increased complexity. However, the inference time remains sufficiently low and enables real-time applications. Figure \ref{fig:xyz} shows an example of the predicted elbow and wrist positions along some arm trajectory, based on FMG measurements. 

To assess data requirements and robustness, we conducted an analysis of model performance across varying numbers of recorded sessions. Each session consists of 20,000 FMG samples, and the FMG device was repositioned between sessions to ensure data diversity. To evaluate the impact of dataset size, the Transformer-based model was trained 15 times for each $K\in\{1,...,48\}$ value, using randomly sampled sessions from the dataset. The resulting position errors are presented in Figure \ref{fig:error_data}. The error is reduced with the increase of session data. Notably, elbow and wrist estimation errors approach their final levels with approximately half the total sessions, indicating rapid convergence. Furthermore, the standard deviation across 15 trials decreases with more sessions, suggesting improved model robustness. The observed performance improvements underscore the significance of data variability introduced by repeated device placements, demonstrating the model's enhanced robustness to variations in positioning and tightening forces.


\subsection{Feature Importance}

\begin{figure}[h]
    \centering
    \includegraphics[width=\linewidth]{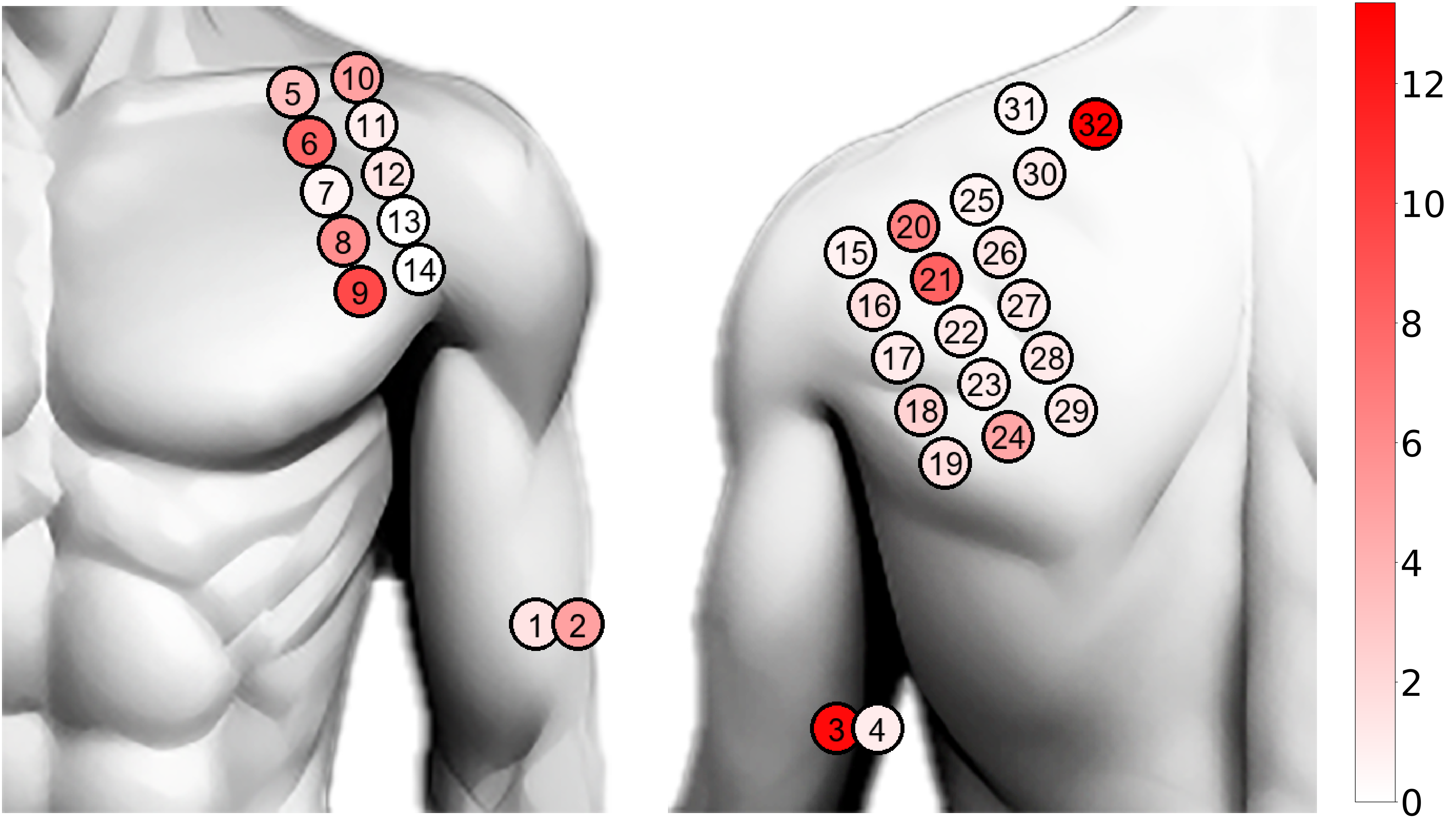}
    \vspace{-0.8cm}
    \caption{Sensor locations and heatmap of feature importance scores for the FSR sensors on the FMG device.}
    \label{fig:hands_sensors}
    \vspace{-0.3cm}
\end{figure}

To assess the relative importance of each FSR sensor, we employed permutation feature importance \cite{Yang2010}, a common technique for evaluating feature contributions in learned models. By randomly shuffling the values of individual sensor readings and measuring the resulting increase in prediction error, we can quantify the importance of each sensor in the model's decision-making process. The score is the error computed according to 
\begin{equation}
    E_i=\frac{e_i-e}{e} \times 100\%,
\end{equation}
where $e$ is the mean error of the non-permuted model and $e_i$ is the mean error when feature $i$ is permuted. Table \ref{tb:score} presents the feature importance scores for each FSR sensor on the user's arm, averaged across wrist and elbow position errors. Sensor placements and importance score heatmap are illustrated in Figure \ref{fig:hands_sensors}. Out of the 32 sensors, 11 were found to be particularly influential, with importance scores exceeding 2.5\%. 

\begin{table}[h]
\vspace{-0.3cm}
\centering
\caption{Feature importance scores of the FSR sensors}
\vspace{-0.2cm}
\label{tb:score}
\begin{adjustbox}{width=\linewidth}
\begin{tabular}{cccccccc}
\toprule
Sensor & Score & Sensor & Score & Sensor & Score & Sensor & Score \\
Index & (\%) & Index & (\%) & Index & (\%) & Index & (\%) \\
\midrule
1 & 1.39 & 9    & 9.52 & 17   & 0.99 & 25   & 0.62  \\
2 & 4.93 & 10   & 4.94 & 18   & 2.36 & 26   & 1.18  \\
3 & 12.82& 11   & 0.89 & 19   & 1.63 & 27   & 1.07  \\
4 & 0.99 & 12   & 1.25 & 20   & 6.49 & 28   & 1.16  \\
5 & 3.37 & 13   & 0.07 & 21   & 8.23 & 29   & 1.11  \\
6 & 7.88 & 14   & 0.01 & 22   & 1.07 & 30   & 0.91  \\
7 & 0.5  & 15   & 0.66 & 23   & 0.94 & 31   & 0.49  \\
8 & 5.88 & 16   & 1.58 & 24   & 4.65 & 32   & 13.35 \\             
\bottomrule
\end{tabular}
\end{adjustbox}
\vspace{-0.5cm}
\end{table}

The above results suggest that some sensors can be removed without a significant compromise in accuracy. To further evaluate this, we progressively remove sensors with importance scores below specific thresholds and evaluate the accuracy of a newly trained model with the retained sensors. In the results presented in Table \ref{tb:score_thresh}, eliminating low-scoring sensors is shown to insignificantly affect accuracy, and moderately high thresholds preserve acceptable performance while reducing the overall sensor count. Hence, a more compact FMG device can be developed, potentially lowering both hardware costs and system complexity.

\begin{table}[h]
\vspace{-0.3cm}
\centering
\caption{Sensor reduction impact on accuracy}
\vspace{-0.3cm}
\label{tb:score_thresh}
\begin{tabular}{cccc}
\toprule
 Importance & Sensors & \multicolumn{2}{c}{Error (mm)} \\\cmidrule{3-4}
 threshold (\%) & retained & Wrist & Elbow \\
\midrule
0.5  & 28 & 121 $\pm$ 66 & 59 $\pm$ 34 \\
1.0  & 21 & 127 $\pm$ 67 & 64 $\pm$ 36 \\
1.5  & 14 & 138 $\pm$ 78 & 71 $\pm$ 41 \\
2.0  & 12 & 140 $\pm$ 77 & 72 $\pm$ 39 \\
2.5  & 11 & 147 $\pm$ 81 & 76 $\pm$ 42 \\
\bottomrule
\end{tabular}
\vspace{-0.5cm}
\end{table}



\subsection{Model Transfer}

To evaluate the model's generalization capabilities, we tested it on three new participants with different Chest Circumferences (CC), Shoulder Widths (SW) and Upper-Arm Circumferences (UAC), who were not included in the original training dataset. We evaluate the zero-shot (ZS) transfer to the new participant and fine-tuning (FT) with a limited amount of data. For the FT, each new participant contributed 20,000 samples over two sessions and an additional 10,000 labeled samples for testing, with the same procedure outlined in Section \ref{sec:data_collection}. FT for a new participant is performed by only retraining the model's last layers of the FC networks, with a learning rate of 
$10^{-5}$.
Table \ref{tb:new_users} summarizes the results along with anthropometric measures of all participants including the one used for training. While the model's accuracy decreased slightly when tested on new participants in ZS with different anthropometric measurements, the results demonstrate its ability to generalize and perform effectively. To improve transferability, a limited amount of additional data from the new participant is shown to significantly improve accuracy.

\begin{table*}[h]
\centering
\begin{minipage}{0.55\textwidth}
\centering
\caption{Accuracy of model transfer to new users}
\label{tb:new_users}
\begin{tabular}{ccccccccc}
    \toprule
    \multirow{2}{*}{User} & CC & SW & UAC & \multicolumn{5}{c}{Position error (mm)} \\\cmidrule{5-9}
    & (mm) & (mm) & (mm) & \multicolumn{2}{c}{Elbow} && \multicolumn{2}{c}{Wrist} \\\cmidrule{5-6}\cmidrule{8-9}
    & & & & ZS & FT && ZS & FT \\\midrule
Train   & 98 & 44 & 28 & 60 $\pm$ 60 & - && 120 $\pm$ 33 & - \\
1       & 99 & 46 & 30.5 & 100 $\pm$ 57 & 82 $\pm$ 34 && 190 $\pm$ 90 & 139 $\pm$ 63\\
2       & 92 & 44 & 25.5 & 93 $\pm$ 41 & 59 $\pm$ 31 && 178 $\pm$ 73 & 120 $\pm$ 62 \\ 
3  & 96 & 50 & 34.5 & 99 $\pm$ 42 & 69 $\pm$ 37 && 178 $\pm$ 79 & 122 $\pm$ 65 \\ 
\bottomrule
\end{tabular}
\end{minipage}
\hfill
\begin{minipage}{0.35\textwidth}
\centering
\caption{HRI demonstration success rates}
\label{tb:demo}
\begin{tabular}{lcc}
\toprule
               & \multicolumn{2}{c}{Session} \\\cmidrule{2-3}
               & ~~~~~1~~~~~              & ~~~~~2~~~~~          \\\midrule
Success rate~~~~~~~   & 84\%           & 90\%          \\
False positive~~~~~ & 5\%            & 10\%          \\
False negative~~~~~ & 11\%           & 0\%         \\
\bottomrule
\end{tabular}
\end{minipage}
\vspace{-0.6cm}
\end{table*}


\begin{figure*}[h]
    \vspace{0.5cm}
    \centering
    \includegraphics[width=\linewidth]{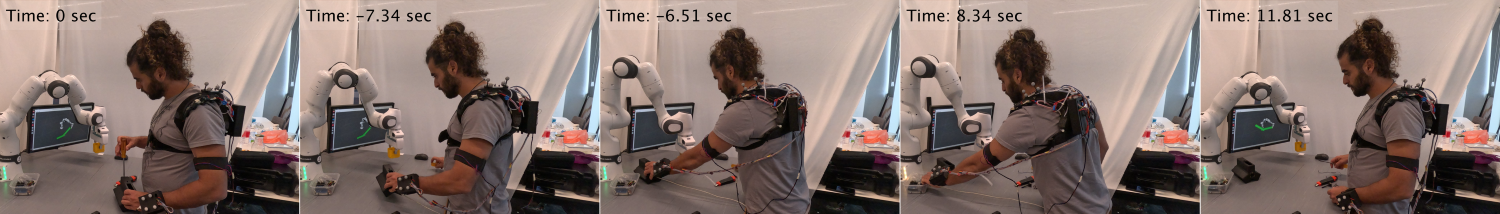}
    \vspace{-0.6cm}
    \caption{A user wearing an FMG device interrupts the robot's motion. The robot pauses its motion based on FMG-based state estimation and resumes only when the user's arm is no longer in the robot's path.}
    \label{fig:demo1}
\end{figure*}

\begin{figure*}[h]
    \centering
    \includegraphics[width=\linewidth]{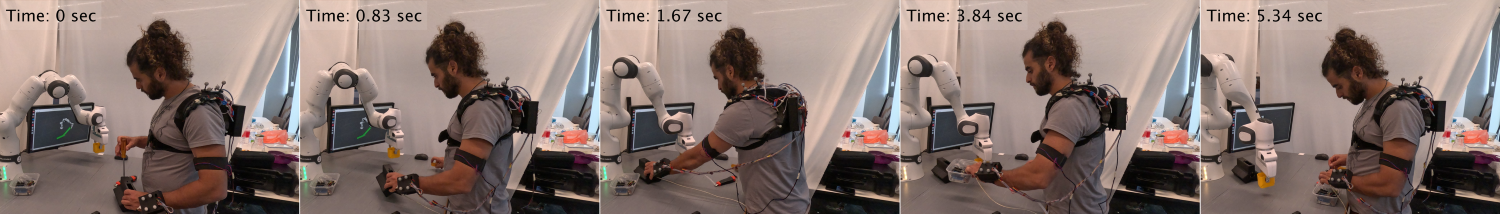}
    \vspace{-0.6cm}
    \caption{A user wearing an FMG device puts down an object and picks up a new one in a shared workspace, while crossing the path of the robot. The robot pauses its motion based on FMG-based state estimation and resumes only when the user's arm is no longer in the robot's path.}
    \label{fig:demo2}
\end{figure*}

\begin{figure*}[h]
    \centering
    \includegraphics[width=\linewidth]{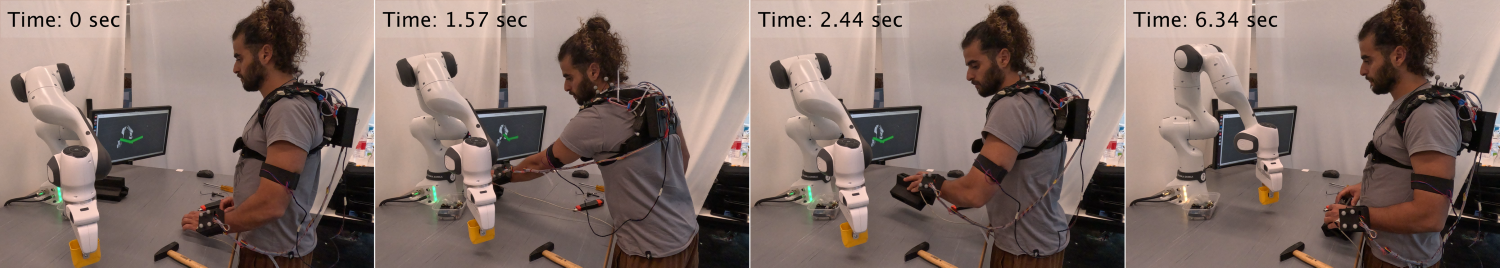}
    \vspace{-0.6cm}
    \caption{A user wearing an FMG device picks up an object in a shared workspace, while crossing the path of the robot. The robot pauses its motion based on FMG-based state estimation and resumes only when the user's arm is no longer in the robot's path.}
    \label{fig:demo3}
    \vspace{-0.5cm}
\end{figure*}



\subsection{Real-time HRI Demonstration}

We have conducted an HRI demonstration in a shared workspace scenario. The Franka collaborative robotic arm was positioned $0.72$ meters from a user on a workstation table as seen in Figure \ref{fig:Cover}. The robot was programmed to move in a pre-defined pick-and-place path across the workstation. No cameras were used to estimate the pose of the user. The nominal position of the human torso is assumed to be known. By observing the state of the human arm solely based on FMG measurements, the robot can instantly halt its motion to avoid interfering with the user's arm path. If a potential collision with the user arm is detected, the robot will stop its motion until its path is clear. Minimal collision clearance was defined to be 150 mm based on the accuracy of the model.

Two experimental sessions were conducted. In the first session, the user deliberately interfered the motion of the robot by blocking its path. An example of such trial is demonstrated in Figure \ref{fig:demo1}. 
In the second session, the user naturally performed various tasks, such as picking up and placing tools and other objects scattered on the table (Figures \ref{fig:demo2}-\ref{fig:demo3}), while blocking the path of the robot. 
Each of the sessions consisted of 60 trials. Table \ref{tb:demo} presents the success rate for the two sessions, including metrics for false positives with failure to stop and false negatives with unnecessary stops. Reasons for false positives and negatives may stem from momentary large errors or out-of-distribution motions, such as high accelerations, which caused the robot to incorrectly estimate a clear path or a potential collision, respectively. Nevertheless, the majority of trials resulted in true positives, demonstrating the potential of this approach for natural HRI, without visual perception.

\section{Conclusions}

We investigated the use of a low-cost wearable FMG device for human arm pose estimation in Human-Robot Interaction (HRI) scenarios. This device, worn on the user's shoulder, incorporates 32 FSR sensors. Using a Transformer-based model, FMG measurements are mapped to the positions of the elbow and wrist, with respect to the shoulder. Our approach exhibited the ability to provide robust, environment-agnostic and fairly accurate arm pose estimations. Unlike common IMU solutions for arm pose estimation, the FMG device can provide drift-free and absolute estimations without the need for frequent calibrations. The trained model was also shown to be transferable to new users with slight decline in accuracy. Furthermore, a set of HRI experiments has demonstrated the robot's ability to avoid collisions using the estimated arm poses in a shared workspace. Generally, the FMG device has the potential to enhance various HRI applications, including medical procedures, factory work and assistive technologies.

While the estimation model was shown to be able to fairly generalize to new users, future work should consider making the model more robust to new users. One approach to enhance the model's generalization capabilities is to train it on a diverse dataset encompassing data from multiple users, as demonstrated in \cite{Bamani2022}. Alternatively, future work could involve customizing both the FMG device and the model to individual users. This could include 3D scanning a user's body to create a custom-fitted device, ensuring optimal sensor placement and data quality. Then, the model can be customized for the user through fine-tuning with a limited amount of additional data, as in \cite{Mizrahi2024}. 

To further enhance accuracy, future work could extend the model to estimate limb orientation. This additional information could help mitigate error propagation and refine wrist position estimations. Furthermore, limb orientation estimates could be valuable for tasks such as human-robot hand-over interactions. These suggestions have the potential to also mitigate false positive and negative readings, while future work could also expand the training dataset to include complex and high-velocity motions, thereby enabling the system to handle rapid and less common movements effectively. Unlike EMG, FMG can be placed-on or integrated into the fabric of clothing, enhancing comfort and user experience. Hence, further work can integrate FSR sensors into wearable textiles and open up new possibilities for advanced and more spontaneous HRI. Future work could also explore the integration of FMG with vision to create a robust hybrid system capable of handling diverse environmental conditions. FMG can provide robust signals unaffected by visual uncertainties, while visual perception can enhance spatial awareness and localize general human pose within the environment.

\bibliographystyle{IEEEtran}
\bibliography{ref}

\end{document}